\documentclass{article}
\usepackage{ijcai21}

\usepackage{times}
\usepackage{soul}
\usepackage{url}
\usepackage[hidelinks]{hyperref}
\usepackage[utf8]{inputenc}
\usepackage[small]{caption}
\usepackage{graphicx}
\usepackage{amsmath}
\usepackage{amsthm}
\usepackage{booktabs}
\usepackage{algorithm}
\usepackage{algorithmic}
\usepackage{overpic}
\usepackage{multirow}
\usepackage{pifont}
\newcommand{\cmark}{\ding{51}}%
\usepackage[symbol]{footmisc}

\newcommand{\x}{{\mathbf x}}
\newcommand{\y}{{\mathbf y}}
\newcommand{\X}{{\mathcal X}}
\newcommand{\Y}{{\mathcal Y}}
\newcommand{\yref}{\y_{\text{ref}}}
\newcommand{\cref}{\c_{\text{ref}}}
\renewcommand{\c}{{\mathbf c}}

\usepackage[dvipsnames]{xcolor}

\title{Enhance Images as You Like with Unpaired Learning}

\author{
Xiaopeng Sun$^{1}$\footnote{Equal contribution.}
\and
Muxingzi Li$^{2}$\footnotemark[1]\footnote{Corresponding author.}\and
Tianyu He$^{2}$\And
Lubin Fan$^{2}$
\affiliations
$^1$Xidian University\\
$^2$Alibaba Group
\emails
{xpsun@stu.xidian.edu.cn},
{\{muxingzi.lmxz,timhe.hty\}@alibaba-inc.com},
{lubinfan@gmail.com}
}

\begin{document}

\maketitle

\begin{abstract}
Low-light image enhancement exhibits an ill-posed nature, as a given image may have many enhanced versions, yet recent studies focus on building a deterministic mapping from input to an enhanced version. In contrast, we propose a lightweight one-path conditional generative adversarial network (cGAN) to learn a one-to-many relation from low-light to normal-light image space, given only sets of low- and normal-light training images without any correspondence. By formulating this ill-posed problem as a modulation code learning task, our network learns to generate a collection of enhanced images from a given input conditioned on various reference images. Therefore our inference model easily adapts to various user preferences, provided with a few favorable photos from each user. Our model achieves competitive visual and quantitative results on par with fully supervised methods on both noisy and clean datasets, while being 6 to 10 times lighter than state-of-the-art generative adversarial networks (GANs) approaches.
\end{abstract}


\section{Introduction}
Low-light image enhancement is fundamentally an image-to-image translation problem which aims to map low quality inputs to high quality versions. It is a task focusing on improving visual quality of an underexposed image which suffers from poor visibility, low contrast and noise. Recent works typically learn an one-to-one mapping functions from the perspective of paired data~\cite{deepretinex}, learning unpaired features~\cite{jiang2019enlightengan} and brightness constraints~\cite{guo2020zero}. However, this mapping is not necessarily one-to-one, as one may want to generate from one input image multiple enhanced versions with different characteristics (lighting, tone, details etc), and meanwhile one high quality image can correspond to many low quality versions. This ill-posed nature indicates the unsuitability of paired supervision with one-to-one mapping assumption in image enhancement tasks, which exactly motivates our work. In this work, we define the task as improving the color, brightness and contrast of the input image conditioned on a given reference image. The use of conditioning enables one-to-many learning. 

\begin{figure}[!t]
    \centering
    \vspace{3pt}\hspace{-14pt}
    \begin{overpic}[trim=0cm 0cm 0cm 0cm,clip,width=0.97\linewidth,grid=false]{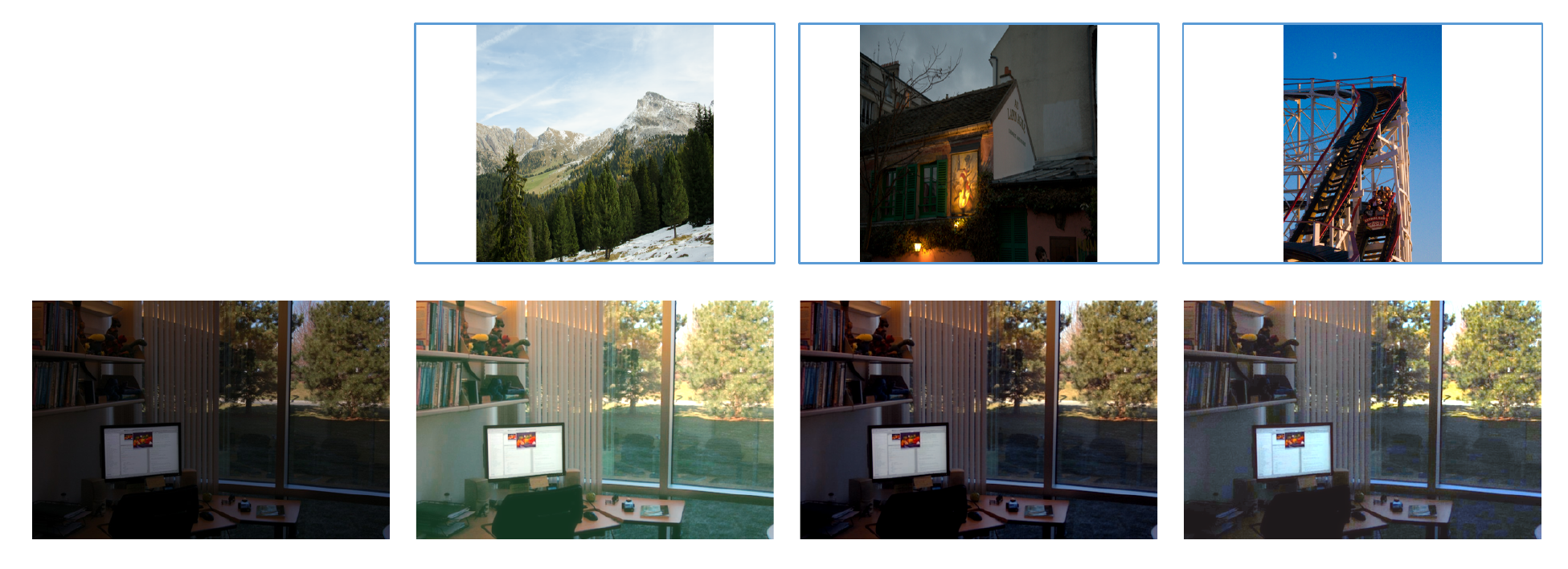}

    \put(9.5,-2){\footnotesize Input}
    \put(58,-2){\footnotesize Output}
    \put(7.5,31){\footnotesize Reference}
    
    \end{overpic}
    \vspace{-0pt}
    \caption{One-to-many Image Enhancement. We show an input low-light image at the bottom left and a set of normal-light reference images with different style on the first row. We then show our stylized results with enhanced light condition.}
    \label{fig:eg_one2many}
\end{figure}

In this paper, we focus on learning an one-to-many mapping model without paired training samples. Concretely, as illustrated in Figure~\ref{fig:eg_one2many}, we are able to translate a given low-light image to a normal-light one conditioned on the reference image (e.g., user preference on the image brightness, contrast, etc.). The conditional enhancement procedure is conducted by a U-Net Translator and a Modulation Code Generator (MCG). 
Specifically, the MCG generates a modulation code that fuses the characteristics of the learned features of both the low-light input image and the reference image. Meanwhile, the U-Net Translator performs conditional translation on the input low-light image with the assistance of our proposed Pixel-wise Self-Modulation (PSM) and Channel-wise Conditional-Modulation (CCM). The PSM module learns to adjust the mean and variance of the feature of input low-light image on each spatial location,  while CCM is complementary to this operation. It performs channel-wise modulation conditioned on the modulation code generated by the MCG.

To enable unpaired learning, we optimize the model with four objective functions: 1) the idempotence loss that assumes a normal-light image should be mapped to itself when conditioned on itself. 2) the spatial consistency loss that facilitates the generated image to have more spatial coherence with the input. 3) the global color consistency loss that makes the overall color coherent with the input. 4) the GAN loss that tries to make the outputs more realistic.

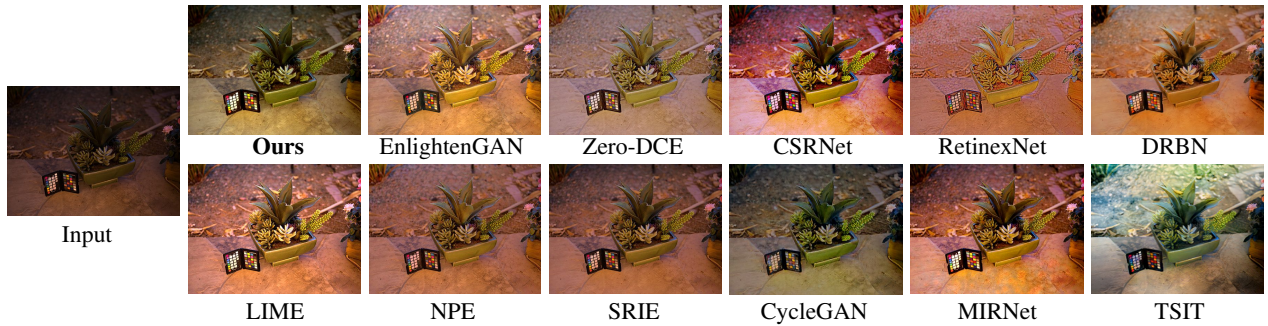
\begin{figure*}[!t]
    \centering
    \begin{overpic}[trim=0cm 4.5cm 0cm 5.5cm,clip,width=0.95\linewidth,grid=false]{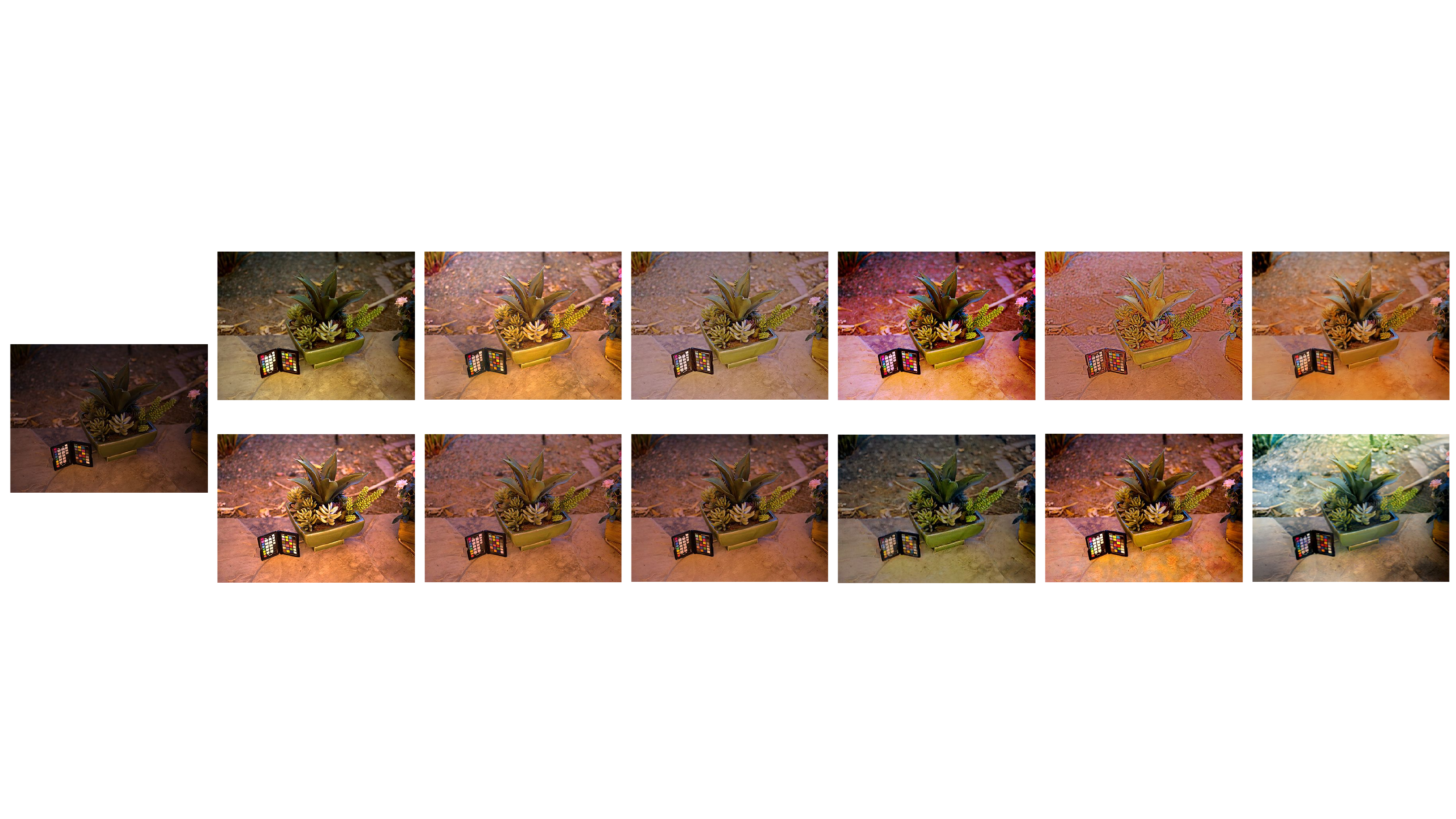}
    \put(5,7){\footnotesize Input}
    \put(20,14){\footnotesize \textbf{Ours}}
    \put(30,14){\small EnlightenGAN}
    \put(46,14){\footnotesize Zero-DCE}
    \put(61,14){\footnotesize CSRNet}
    \put(74,14){\footnotesize RetinexNet}
    \put(90,14){\footnotesize DRBN}
    
    \put(19.5,1){\footnotesize LIME}
    \put(34,1){\footnotesize NPE}
    \put(48,1){\footnotesize SRIE}
    \put(60,1){\footnotesize CycleGAN}
    \put(75.5,1){\footnotesize MIRNet}
    \put(91,1){\footnotesize TSIT}
    \end{overpic}
    \vspace{-10pt}
    \caption{Our method outperforms the state-of-the-art baselines both qualitatively and quantitatively.}
    \label{fig:res_eg1}
\end{figure*}

\section{Related Works}
We review related works in two main categories: image enhancement and conditional image generation. We mainly discuss the literature that addresses the problem of low-light image enhancement in unpaired and unsupervised setting, which is closely related to our setting.

\subsection{Image Enhancement}

\paragraph{Traditional Methods.} 
There are two main categories of methods for low-light image enhancement, Histogram Equalization (HE) based methods~\cite{HE} and Retinex~\cite{retinex}.
For example, LIME~\cite{guo2018lime} searches the maximum value in the RGB channels of the image to estimate the illumination of each pixel then rebuilt the illumination map with a structure prior.  
However, these methods have poor generalization ability, and often result in visible noise for real low-light images.

\begin{figure*}[!t]
    \centering
    \begin{overpic}[trim=0cm 7.85cm 3.75cm 0cm,clip,width=1\linewidth,grid=false]{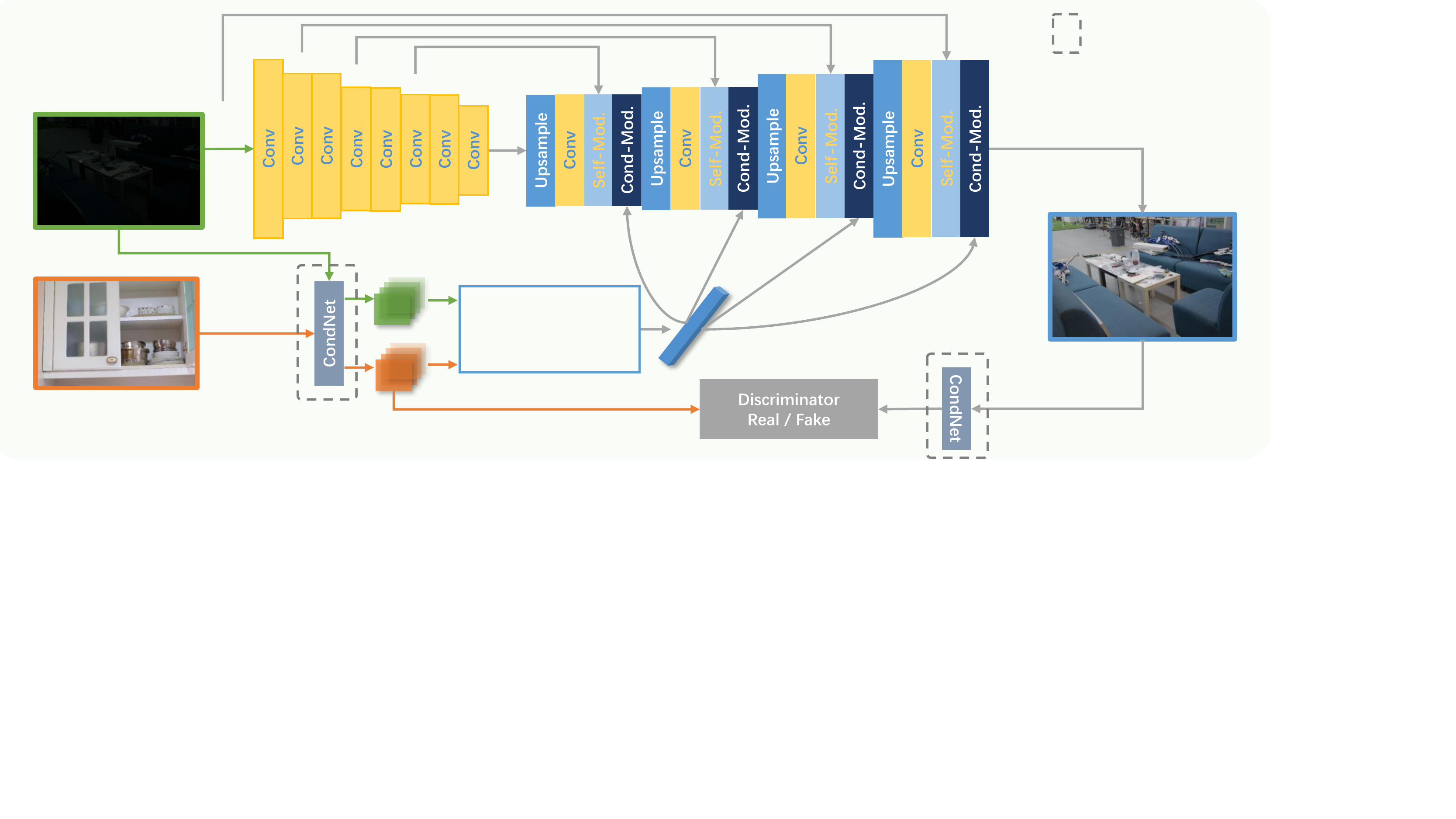}
    \put(2,35){Generator}
    \put(5,29){$X$ Image}
    \put(3,4){Reference Image}
    \put(17,27){$X$}
    \put(21,32.5){\tiny 6}
    \put(23.2,31.5){\tiny 6}
    \put(25,31.5){\tiny 12}
    \put(27.5,30.5){\tiny 12}
    \put(30,30.5){\tiny 24}
    \put(32,29.8){\tiny 24}
    \put(34.5,29.8){\tiny 48}
    \put(36.7,29){\tiny 48}
    \put(44,29.8){\tiny 24}
    \put(52.8,30.5){\tiny 12}
    \put(62,31.5){\tiny 6}
    \put(71,32.5){\tiny 3}
    
    \put(38.3,12.8){Modulation}
    \put(41,10.8){Code}
    \put(39,8.8){Generator}
    \put(90,8){\footnotesize Enhanced}
    \put(91,6){\footnotesize Image}
    
    \put(85,33.6){\footnotesize Weight Sharing}
    \end{overpic}
    \vspace{-21pt}
    \caption{Overview of our Condition GAN for image enhancement. }
    \label{fig:overview}
\end{figure*}
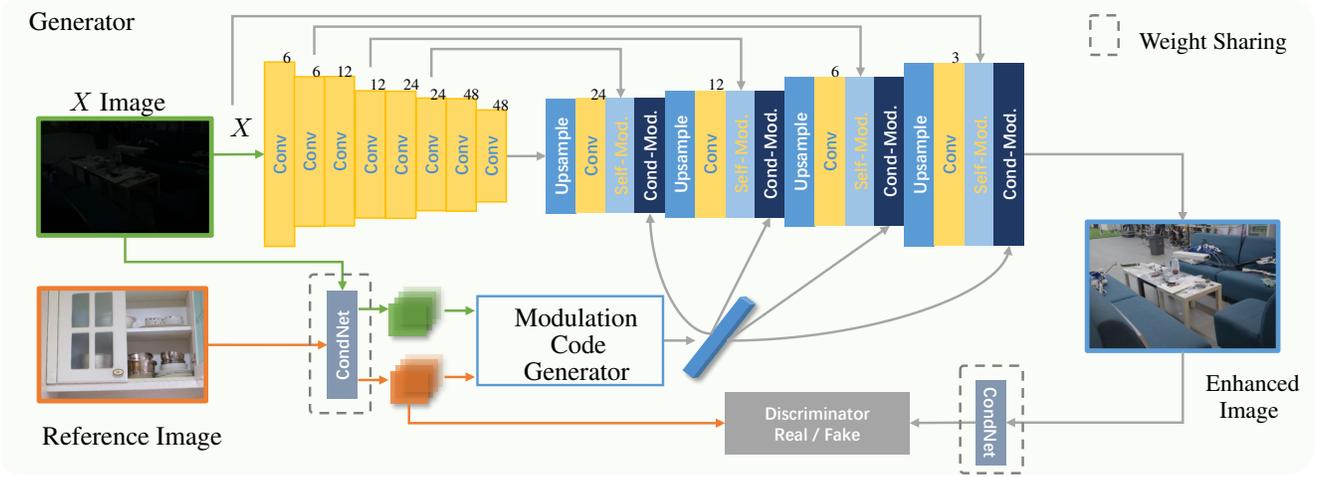

\paragraph{Learning-based Methods with Paired Supervision.} 
Recently, the methods based on deep neural network achieve impressive results on low-light image enhancement. In enhancing natural images, there are many promising methods~\cite{llnet,deepretinex}. And for relight HDR images, HDR-Net~\cite{gharbi2017deep} utilized bilateral grid processing and local affine color transforms. Aiming at enhance raw sensor data, ~\cite{chen2018learning} proposed a “learning to see in the dark” methods that achieves impressive visual results.

\paragraph{Learning-based Methods with Unpaired Supervision.} 
In image translation, there are many excellent works~\cite{iccv2017cyclegan,wang2018high} based on unpaired data. Focusing on image enhancement, several methods are proposed due to the difficulty of obtaining paired data in real scenes. ~\cite{yang2020fidelity} train a deep recursive band network with paired-unpaired images. \cite{jiang2019enlightengan} propose EnlightenGAN that can be trained without low/normal-light image pairs. Zero-DCE~\cite{guo2020zero} estimates the pixel-wise and the high-order curves for dynamic range adjustment of a low-light image in an unsupervised way. These methods are one-to-one mapping of low light images to target domain. However, image enhancement is ill-posed and cannot be inverted with a deterministic mapping.
While different from aforementioned works that map low-light images to a single enhanced distribution, we develop a lightweight conditional GAN, that learns a one-to-many relation from low-light to normal-light image space without paired datasets.

\subsection{Conditional Image Generation}
The generative adversarial networks~\cite{goodfellow2014generative} (GANs) employ a discriminator to distinguish the generated images from the real ones. Prior works have conditioned GANs (i.e., cGANs) on discrete labels~\cite{mirza2014conditional}, text~\cite{reed2016generative} or images~\cite{isola2017image}.

Among them, the most related direction to ours is translating an image from one domain to another, conditioned on a given reference image. Along this line, previous works in image style transfer introduce Conditional Instance Normalization (Conditional IN) and Adaptive Instance Normalization (AdaIN) to adjust the mean and the variance of the content input by style-specific parameters~\cite{dumoulin2016learned} or alternatively by directly replacing the mean and the variance with those of the style input~\cite{huang2017arbitrary}. Basically, these normalization-based methods first normalize the features to a normal distribution, then de-normalize them with a learned affine transformation whose parameters inferred from external data. Due to their flexibility, both were successfully adopted in various tasks with paired supervision~\cite{brock2018large,park2019semantic,zhang2020cross}.

Most of conditional image generation works are trained with paired data, e.g., segmentation masks and images. While in this paper, we focus on unpaired conditional image generation, and achieve it with several proposed schemes that are elaborately tailored for image enhancement (see Fig.~\ref{fig:res_eg1} for a qualitative comparison between our method and above mentioned baselines.).

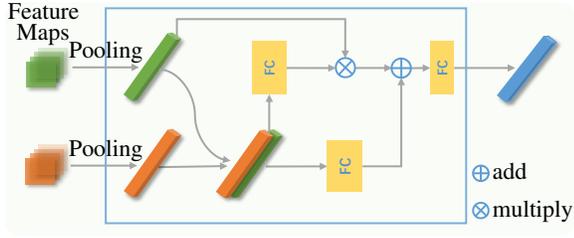
\begin{figure}[!t]
    \centering
    \begin{overpic}[trim=4cm 12.2cm 13.5cm 0cm,clip,width=0.9\linewidth,grid=false]{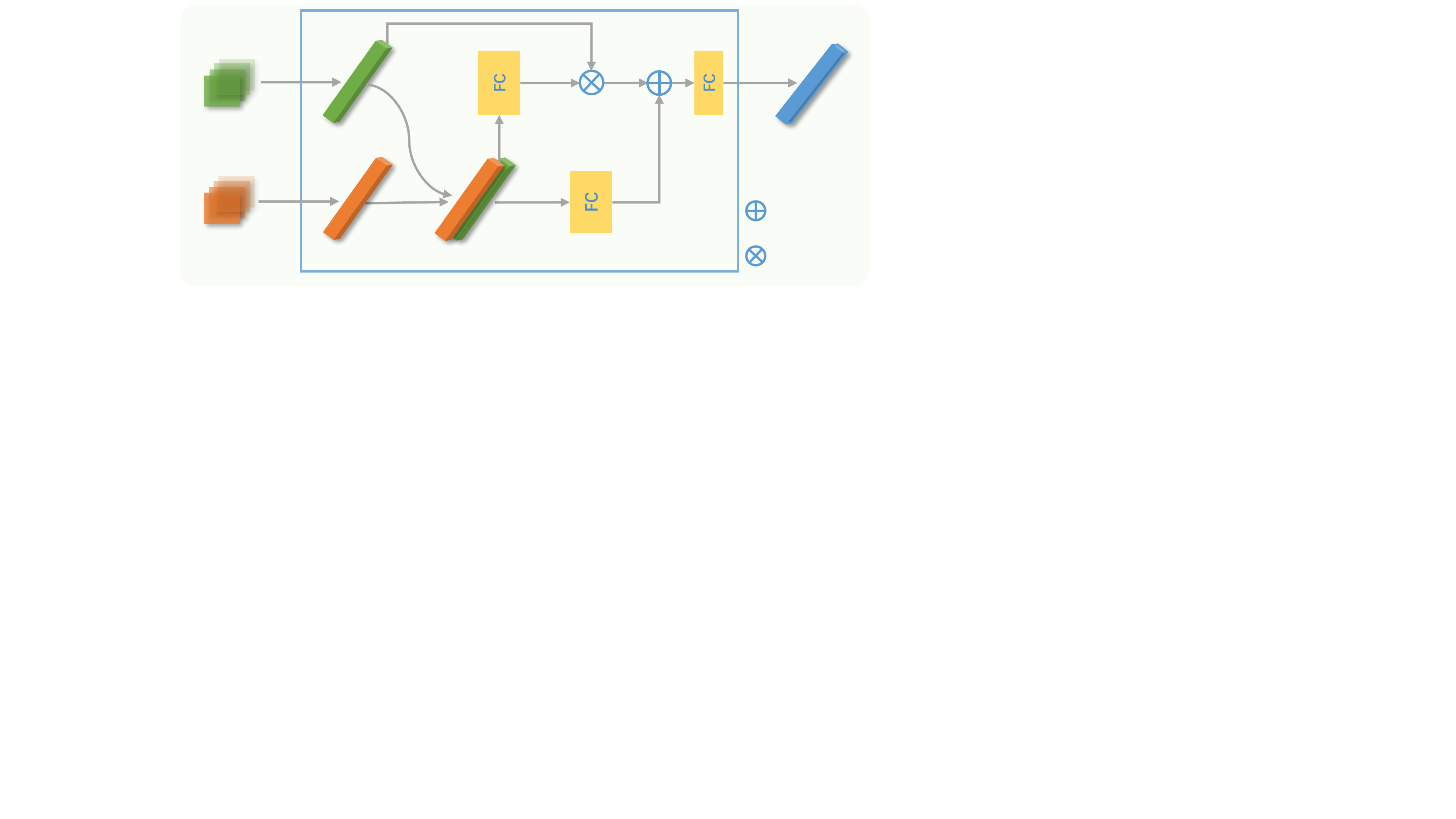}
    \put(12,31.8){\footnotesize Pooling}
    \put(12,14.8){\footnotesize Pooling}
    \put(1.5,38.5){\footnotesize Feature}
    \put(3,35){\footnotesize Maps}
    \put(85,11){\footnotesize add}
    \put(85,4.5){\footnotesize multiply}
    \end{overpic}
    \vspace{-10pt}
    \caption{Modulation Code Generator}
    \label{fig:code}
\end{figure}

\section{Methodology}

\paragraph{Problem Formulation.}
We aim at transferring a given low-light image to its normal-light counterpart according to user's preference without paired training samples. Formally, let $\x$ be an input image in the image space $\X$. We want to construct a conditional mapping $G(\x \,|\,\yref): \x\in\X \to \y\in\Y$ which maps $\x$ to an image $\y$ in the target image space $\Y$ of normal-light images, conditioned on the reference image $\yref$. To this end, we propose to transfer the style information contained in $\yref$ through the form of a modulation code $\cref$  (see Fig.~\ref{fig:code}).

\paragraph{Approach Overview.}
To enhance the input low-light image to a normal-light one conditioned on a reference image, we employ a U-Net Translator that performs conditional translation on the input low-light image. Our U-Net Translator consists of two complementary modules: the Pixel-wise Self-Modulation (PSM) and the Channel-wise Conditional-Modulation (CCM), where both of them are designed to adjust the feature distribution of the low-light input image but from different aspects. Specifically, the PSM is designed to learn modulation parameters from previously upsampled features, while the CCM is designed to learn from the features of both the low-light and the reference image. In particular, the features fed into CCM are generated by our Condition Net (CondNet), which consists of three convolutional layers.
To make the enhanced image natural, we also equip the model with a discriminator that distinguishes the feature outputted by CondNet from low- and normal-light domain. By enabling weight sharing, CondNet is encouraged to learn the difference , between the two image space, seen by both the generator and the discriminator. It prevents discriminator from cheating with discriminating based on other feature that is unrelated to the generator side.

\subsection{Modulation Code Generator}

A typical conditional image generation approach generates modulation parameters (e.g., learned scale and bias for AdaIN~\cite{dumoulin2016learned}) purely based on conditional input, which is the reference image $\yref$ in our setting~\cite{park2019semantic,zhang2020cross}. However, for our unpaired image enhancement, there are several issues of generating modulation parameters from the reference image only: 1) Our goal is to enhance input image itself according to the reference image, which not only fully depends on the reference image but also needs to consider the property of the input image. 2) Since we do not have paired training samples, it is impossible to optimize the model with pair-wise constraint like Mean Square Error. Generating modulation code only conditioned on the reference may allow the network to take shortcuts and cheat on the loss function where low cost is achieved while no valid characteristic is extracted from the reference image. More specifically, the network would learn a \emph{constant bias} while only take the reference image as input. Therefore, we combine the information from both the input image and the reference one to facilitate the learning process of Modulation Code Generator.
Formally, we perform global average pooling on the outputs of CondNet, forming two feature vectors $\x^c$ and $\yref^c$ for the input image $\x$ and the reference image $\yref$ respectively. Our Modulation Code Generator can be formulated as:
\begin{equation}
  \begin{aligned}
    \cref = & \texttt{fc}_\texttt{out} ( \texttt{fc}_\texttt{in} (\cref') \odot \yref^c \oplus \texttt{fc}_\texttt{y} (\cref') ),  \\
    \text{where} \; \; \cref' = & \texttt{concat} (\x^c, \yref^c),
  \end{aligned}
\end{equation}
$\texttt{concat}$ and $\texttt{fc}$ indicate concatenation operation and fully-connected layer respectively, $\odot$ and $\oplus$ denote element-wise multiplication and addition respectively.

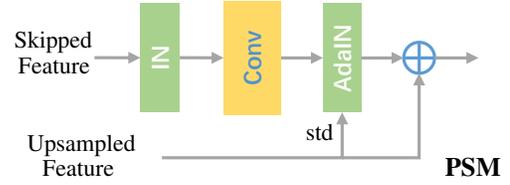
\begin{figure}[!t]
    \centering
    \begin{overpic}[tics=10,trim=0cm 9.2cm 24.8cm 5.7cm,clip,width=0.8\linewidth,grid=false]{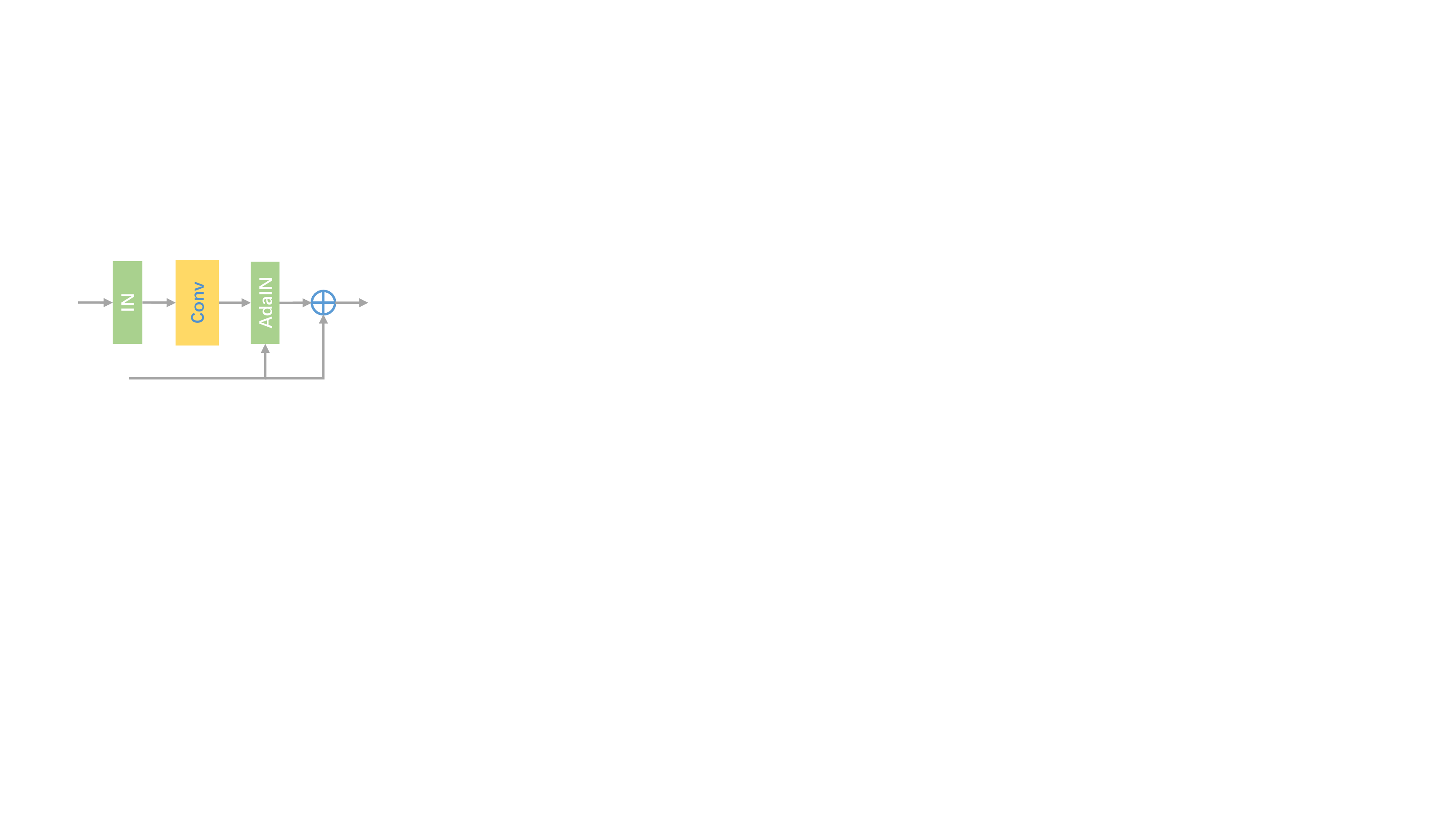}
    \put(88,8){\textbf{PSM}}
    \put(7,13){\footnotesize Upsampled}
    \put(10,8){\footnotesize Feature}
    \put(4.5,33){\footnotesize Skipped}
    \put(5.3,28){\footnotesize Feature}
    \put(61,15){\footnotesize std}

    \end{overpic}
    \vspace{-15pt}
    \caption{Pixel-wise Self-Modulation (PSM) block}
    \label{fig:selfmodulation}
\end{figure}

\subsection{U-Net Translator}
\begin{figure}[!t]
    \centering
    \begin{overpic}[trim=19cm 7.1cm 4cm 4cm,clip,width=0.8\linewidth,grid=false]{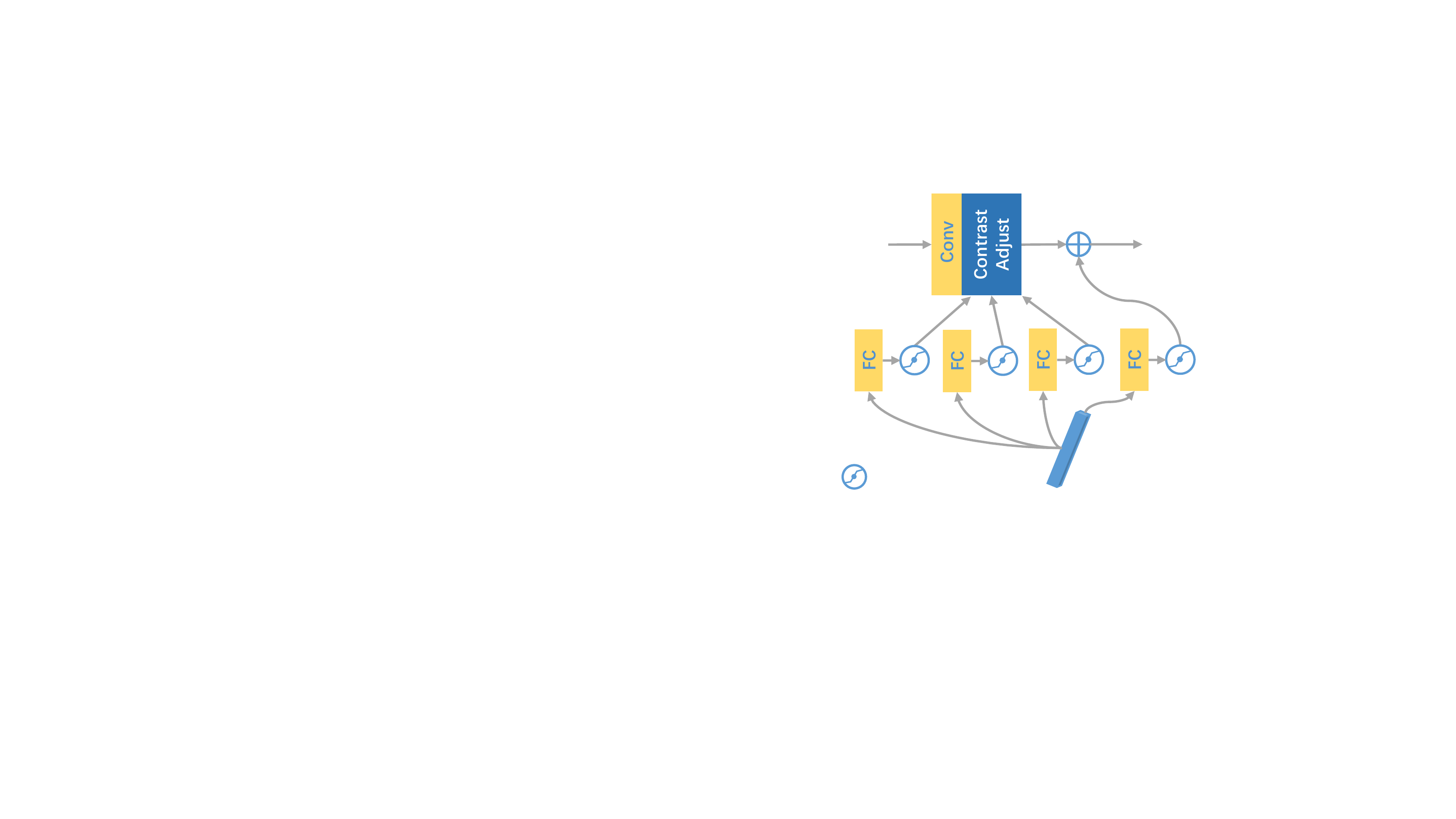}
    \put(88,5){\textbf{CCM}}
    \put(12,7){\footnotesize Activation}
    \put(10,56.5){\footnotesize $X$}
    \put(72,56.5){\footnotesize Output}
    \put(57,13){\footnotesize Modulation Code}
    
    \end{overpic}
    \vspace{-10pt}
    \caption{Channel-wise Conditional-Modulation (CCM) block}
    \label{fig:retouch}
\end{figure}

Our U-Net Translator follows a typical encoder-decoder architecture with skip connections~\cite{ronneberger2015u}, which has strong capability on multi-scale texture preservation~\cite{jiang2019enlightengan}. We tailor the standard U-Net architecture for our task from three aspects: 1) we propose a Pixel-wise Self-Modulation (see Fig.~\ref{fig:selfmodulation}) to match the statistics of the upsampled feature with the skipped feature. 2) we propose a Channel-wise Conditional-Modulation (see Fig.~\ref{fig:retouch}) to polish the feature with the generated modulation code (see Fig.~\ref{fig:code}), which is vital for conditional image enhancement. 3) we remove the original batch normalization layers which destroy the relative distribution across channels, as our Channel-wise Conditional-Modulation is learned more efficiently by seeing the original distributions. We elaborate the two proposed modules as follows.

\paragraph{Pixel-wise Self-Modulation (PSM).}
Commonly, the decoder of the U-Net combines the preceding feature generated from the encoder with the upsampled feature by concatenation or summation. However, in an enhancement setting, the feature generated from the encoder side generally represents the input in the low-light domain. While for the decoder side, we gradually generate normal-light representation for the input. Simply mixing them together without additional adjustment will lead to a domain gap between the encoder and decoder representation. Therefore, instead of a direct concatenation, we propose a Self-modulation Block to adjust the statistics of skipped feature by the upsampled feature from the previous layer. As illustrated in Figure~\ref{fig:selfmodulation}, the skipped feature is first processed by an instance normalization layer and two $3\times3$ convolution layers with leaky ReLU, then fed into AdaIN~\cite{huang2017arbitrary}, whose mean and variance are calculated from the upsampled feature. The goal of our PSM block is to enhance the skipped feature, which consists of multi-scale texture information, in an adaptive way. Intuitively, it can be viewed as modulating the lower-level representation using the higher-level one. Thus, the PSM block is expected to automatically learn to enhance the image while preserving the detailed content. Note that, our PSM block is totally different from~\cite{chen2018self}, where the feature is modulated by some input noise. In contrast, our modulation condition is provided by previous layer, targeting at matching the statistics between the skipped feature and the upsampled one.

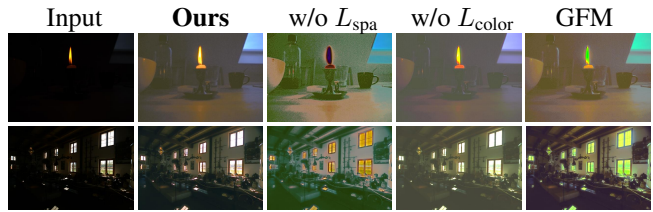
\begin{figure}[!t]
    \centering
    \vspace{10pt}
    \begin{overpic}[trim=0cm 6.0cm 1cm 3cm,clip,width=1\linewidth,grid=false]{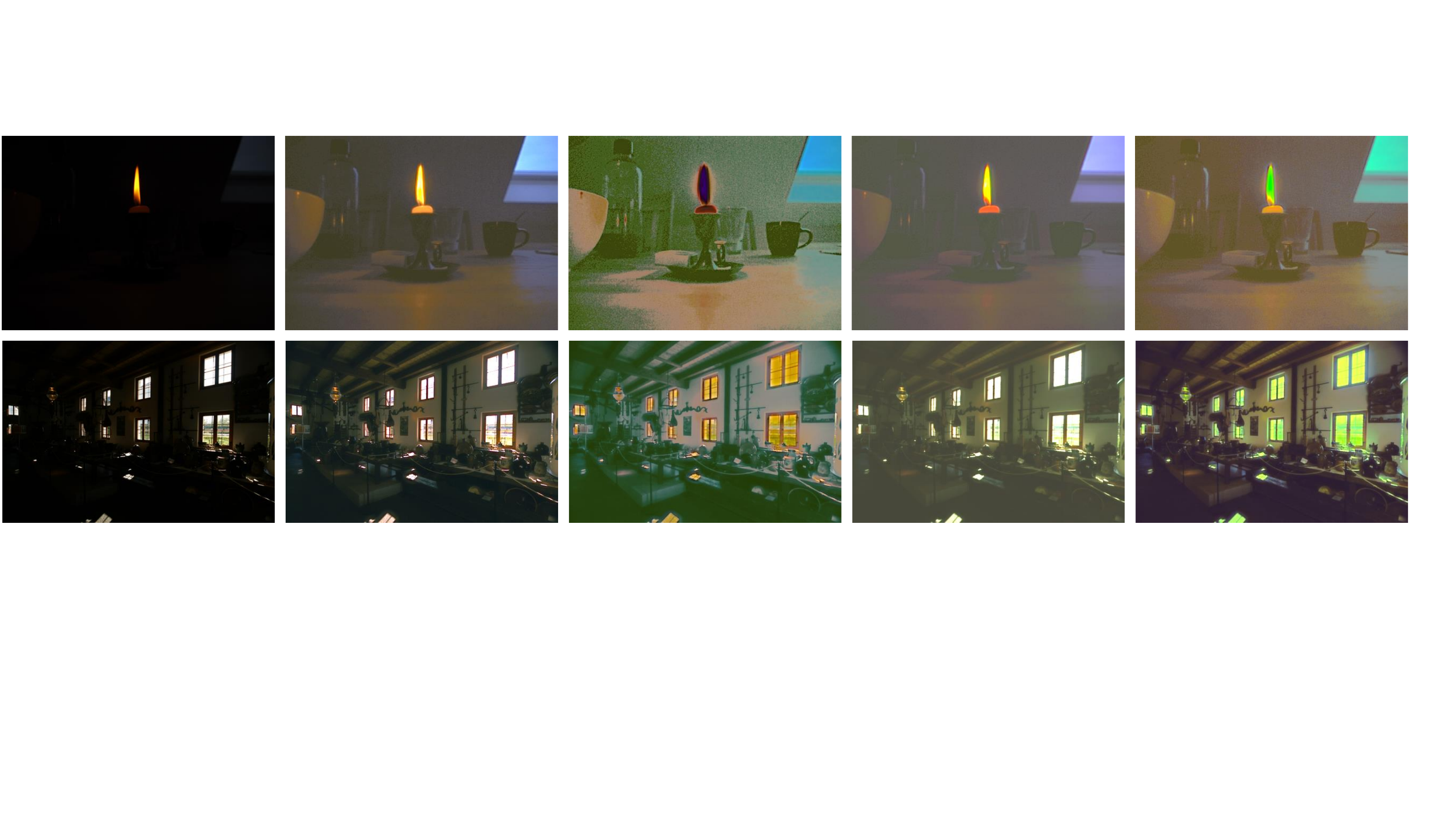}
    \put(6,31.2){Input}
    \put(25.5,31.2){\textbf{Ours}}
    \put(43.5,31.2){w/o $L_{\text{spa}}$}
    \put(62.5,31.2){w/o $L_{\text{color}}$}
    \put(85,31.2){GFM}
    \end{overpic}
    \vspace{-22pt}
    \caption{Ablation study on losses and the CCM block. On column 3-4, we verify the usefulness of our loss functions on two challenge cases of low-light images. In the last column, we show the results of replacing our CCM block with Global Feature Modulation (GFM) followed by ReLU, which justifies the effectiveness of our CCM block.}
    \label{fig:ablation}
\end{figure}

\paragraph{Channel-wise Conditional-Modulation (CCM).}
Our CCM block plays two important roles: 1) it transfers the style of the reference image, which encoded in a modulation code, to the input low-light image. 2) it performs learnable non-linear transformation on the learned feature. Formally, let $\cref$ be the modulation code inferred from $\yref$, as shown in Figure~\ref{fig:retouch}, our CCM block first generates four coefficient vectors $\alpha_{\text{ref}}^1$, $\alpha_{\text{ref}}^2$, $\bar{\alpha}_{\text{ref}}$, and $\beta_{\text{ref}}$ based on the modulation code $\cref$ through two fully-connected layers. Then the retouch operation $m(\cdot)$ can therefore be formulated as:
\begin{equation}
    m(\x)=\left\{
    \begin{aligned}
    \alpha_{\text{ref}}^1 \odot (\x \ominus \bar{\alpha}_{\text{ref}}) \oplus \beta_{\text{ref}} & & \text{if} & \; \; \x > \bar{\alpha}_{\text{ref}} \\
    \alpha_{\text{ref}}^2 \odot (\x \ominus \bar{\alpha}_{\text{ref}}) \oplus \beta_{\text{ref}} & & \text{if} & \; \; \x \leq \bar{\alpha}_{\text{ref}},
    \end{aligned}
    \right.
\label{eq:CCM}
\end{equation}
where $\odot$, $\ominus$ and $\oplus$ indicate element-wise multiplication, subtraction and addition respectively. A typical contrast and brightness adjustment operation in image processing can be formulated as 
$I^{\text{new}} = \alpha I+(1-\alpha)\beta + b$, where $\alpha$ is a scaling factor, $\beta$ is the average intensity value, and $b$ is the brightness adjustment coefficient. which can be simplified as $\alpha I + \gamma$. However, this linear operation has limited effect as compared to the more sophisticated curve adjustment. In previous work, non-linearity is achieved by introducing activation functions like ReLU into the main path \cite{jiang2019enlightengan,he2020conditional}, or by using curve adjustment directly \cite{guo2020zero}. We propose to simulate curve adjustment by compositing the function defined in \eqref{eq:CCM} for multiple times. 


\subsection{Objective Functions}
As we have encoded the brightness and the contrast information into the Condition Net, and further encoded the reference-controlled channel-wise modulation information into the modulation code, our U-Net translator only needs to carry forward content information, and is therefore suitable to take as input images from both low-light and normal-light image space. Our training involves feeding images from both space to the translator. We propose to use two non-reference pixel-wise losses, i.e.~an idempotence loss and a spatial consistency loss, within the target image space, together with a global color consistency loss and a GAN loss to enable unpaired learning in U-Net Translator. 

\paragraph{Idempotence Loss.} 
The idempotence loss requires that a normal-light image should be mapped to itself when conditioned on itself. Let $\y$ be an image sampled from the normal light space, the loss is defined as
\begin{equation}\label{eq:idem}
L_{\text{idem}} = \big\Vert G(\y\,|\,\y) - \y \big\Vert_{1}
\end{equation}

\paragraph{Spatial Consistency Loss.} 
We adopt the spatial consistency loss~\cite{guo2020zero} between the generated reference, which encourages spatial coherence between two images in the form of consistent gradient variation in the local neighborhood. Let $\y_{1}, \y_{2}$ be two images sampled from the normal light space, the loss is given as 
\begin{equation}\label{eq:spa}
L_{\text{spa}} = \big\Vert \nabla G\big(\y_{1}\,|\,\y_{2}\big) - \nabla\y_{1}\big\Vert_{1}
\end{equation}

\paragraph{Global Color Consistency Loss.} 
The relative strength of each color channel of the enhanced output should not deviate significantly from the input. To this end, We propose a global color consistency loss which prevents unrealistic color tone shift. Let $I^{c}(\cdot)$ denote the average intensity value of channel $c$ of an layer-normalized image, we define the loss as
\begin{equation}\label{eq:color}
L_{\text{color}} = \sum_{c\in\{R,G,B\}}\Big(I^{c}\big(G(\x\,|\,\yref)\big) -I^{c}\big(\x\big) \Big)^{2}
\end{equation}

\paragraph{GAN Loss.} 
We slightly modify the standard GAN loss by letting the discriminator see two types of fake images, one generated from an low-light input $\x$ and the other generated from a normal-light input $\y$, both conditioned on a reference normal-light image $\yref$. Overall, the loss is written as 
\begin{equation}
\label{eq:ganloss}
  \begin{aligned}
L_{\text{GAN}} = &\phantom{(1-\lambda)}\mathbf{E}_{\y\sim \Y}\big[\log\big(D(\y)\big)\big]+ \\
        &\hspace{25pt}\lambda\mathbf{E}_{\x}\big[\log\big(1-D\big(G(\x\,|\,\yref)\big)\big)\big] + \\
        &(1-\lambda)\mathbf{E}_{\y,\yref\sim \Y}\big[\log\big(1-D\big(G(\y\,|\,\yref)\big)\big)\big]
 \end{aligned}
\end{equation}

Overall, our final loss function is a combination of each individual constraint:
\begin{equation}
\label{eq:sumloss}
    L_{\text{total}} = L_{\text{idem}} + L_{\text{spa}} + L_{\text{color}} + \alpha L_{\text{GAN}}.
\end{equation}

\setlength{\tabcolsep}{1.2em}
\begin{table*}[!t]
\centering
\begin{tabular}{|l|l|c|c|c|c|} 
\hline
\multicolumn{2}{|l|}{Method $\backslash$ Metric}  & Unpaired & Conditional & LOL-690 & FiveK \\ \hline\hline
\multicolumn{2}{|l|}{SRIE} &  &  & $15.35\,\backslash\,0.559\,\backslash\,7.4022$ & $16.90\,\backslash\,0.750\,\backslash\,4.1352$ \\
\multicolumn{2}{|l|}{LIME} &  &  & $17.97\,\backslash\,0.512\,\backslash\,8.2972$ & $16.67\,\backslash\,0.772\,\backslash\,3.7043$ \\ 
\multicolumn{2}{|l|}{NPE} &  &  & $17.62\,\backslash\,0.481\,\backslash\,8.5105$ & $15.60\,\backslash\,0.736\,\backslash\,3.6475$ \\
\multicolumn{2}{|l|}{RetinexNet} &  &  & $16.17\,\backslash\,0.420\,\backslash\,9.2652$ & $11.89\,\backslash\,0.644\,\backslash\,4.4298$ \\
\multicolumn{2}{|l|}{DRBN} &  &  & $18.71\,\backslash\,0.784\,\backslash\,4.5612$ & $15.07\,\backslash\,0.562\,\backslash\,7.1623$ \\
\multicolumn{2}{|l|}{CSRNet} &  &  & $15.69\,\backslash\,0.408\,\backslash\,8.1343$ & $23.68\,\backslash\,0.896\,\backslash\,3.7492$ \\
\multicolumn{2}{|l|}{EnlightenGAN} & \cmark &  & $18.89\,\backslash\,0.692\,\backslash\,5.0857 $  & $15.47\,\backslash\,0.734\,\backslash\,3.7616$ \\
\multicolumn{2}{|l|}{Zero-DCE} & \cmark &  & $18.47\,\backslash\,0.598\,\backslash\,7.8224$  &    $13.01\,\backslash\,0.557\,\backslash\,7.3117$ \\
\multicolumn{2}{|l|}{CycleGAN} & \cmark &  & $17.42\,\backslash\,0.576\,\backslash\,4.0663$  &    $17.04\,\backslash\,0.681\,\backslash\,4.8327$ \\
\multicolumn{2}{|l|}{TSIT} &  & \cmark & $13.14\,\backslash\,0.533\,\backslash\,5.5965$  &    $14.35\,\backslash\,0.638\,\backslash\,5.3926$ \\
\multicolumn{2}{|l|}{MIRNet} & \cmark &  & $12.90\,\backslash\,0.432\,\backslash\,4.2501$  &    $19.36\,\backslash\,0.806\,\backslash\,3.9225$ \\
\hline\hline
    & Min. &                    &                  & $12.24\,\backslash\,0.609\,\backslash\,-$  & $11.97\,\backslash\,0.655\,\backslash\,-$ \\
\textbf{Ours} & Avg. & \multirow{2}{*}{\cmark} & \multirow{2}{*}{\cmark} & $17.00\,\backslash\,0.671\,\backslash\,-$ & $17.37\,\backslash\,0.750\,\backslash\,-$ \\
    & Max. &                    &                    & $22.45\,\backslash\,0.732\,\backslash\,4.0733$ & $20.87\,\backslash\,0.797\,\backslash\,4.0305$\\ \hline
\end{tabular}
\caption{PSNR($\uparrow$) \textbackslash ~SSIM($\uparrow$) \textbackslash ~NIQE($\downarrow$) metrics on the paired test set of datasets LOL~\protect\cite{deepretinex} starting from image \#690, and FiveK~\protect\cite{fivek2011}. The arrow after each metric indicates whether a larger or a smaller value is better. As our method generates a distribution of output images given a set of reference images, we report the minimum, average and maximum values of PSNR and SSIM. As NIQE is a no ground-truth quality metric, we can thus select the reference which gives the best NIQE, from the reference set.}
\label{tb:psnr_ssim}
\end{table*}

\section{Experiments}

\subsection{Dataset and Implementation Details}
One of the main advantages of our unpaired setting for image enhancement is that we utilize a much larger collection of low-light and normal-light images without imposing given correspondences between the images, which is not the case for the methods designed on paired and fully supervised setting.
Thereby, we assemble images from three different datasets~\cite{deepretinex,fivek2011,exdark2019} and ignore the paired information in each individual dataset if there is any, which leads to a larger and more diverse dataset that consists of $983$ low-light and $5576$ normal-light images. We follow the same practice of previous work\cite{yang2020fidelity} to use part of the LOL dataset\cite{deepretinex} for training, and leaving the other part for testing. We then train our network on this unpaired dataset and compare to other methods with their pretrained models.

We implement our network with PyTorch on a Tesla GPU. Our network has $891,527$ parameters in total including the discriminator, leading to almost $10$ times reduction in size as compared to EnlightenGAN~\cite{jiang2019enlightengan} with $8,636,675$ parameters. The weights of each layer are initialized with random values sampled from a Gaussian with 0 mean and 1 standard deviation. We adopt Adam optimizer with default parameters and with learning rate set to $5\times 10^{-5}$. We set the loss weight $\lambda$ in Eq.~\eqref{eq:ganloss} to $0.9$, and $\alpha$ in Eq.~\eqref{eq:sumloss} to $0.05$ in all the tests. 
Our code can be found at \url{https://github.com/sxpro/ImageEnhance_cGAN}.

\subsection{Ablation Study}
We demonstrate the effectiveness of our choice of losses and the CCM block via ablation studies. We do not ablate the idempotent loss $L_{\text{idem}}$ as it is the only loss that enforces content consistency in our setting, meaning that the generator would produce almost arbitrary results in absence of $L_{\text{idem}}$. 
\paragraph{Contrast adjustment module.} Our Channel-wise Conditional-Modulation (CCM) is designed to transfer the \emph{style} of the reference image which can particularly capture the contrast and the brightness of the reference image. An alternative design choice would be adopting GFM followed by leaky ReLU as used in~\cite{he2020conditional}. However, this can result in non-realistic color tones as shown in the last column in Fig.~\ref{fig:ablation}. 
On the contrary, our proposed CCM component can properly capture the style feature of the reference image and smoothly transfer it to the output.

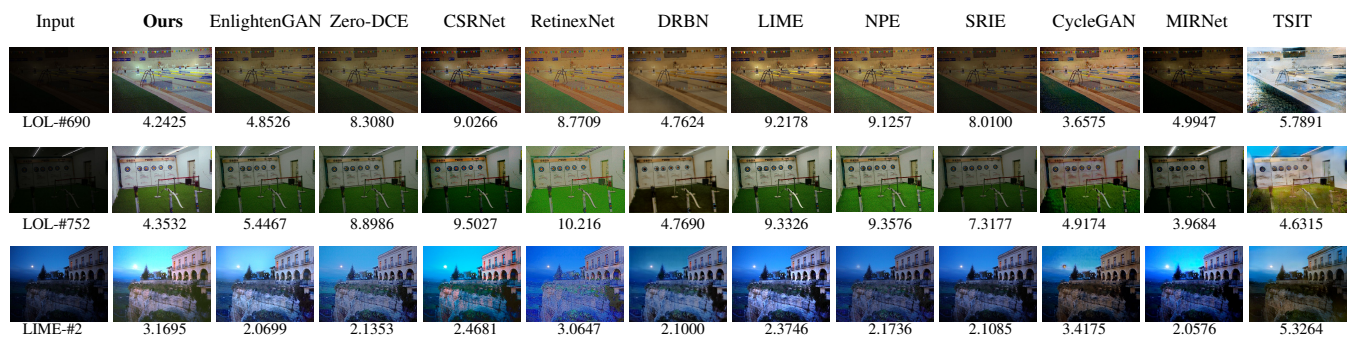
\begin{figure*}[!t]
    \centering
    \vspace{6pt}
    \begin{overpic}[trim=0.0cm 0.0cm 0.0cm 10cm,clip,width=1\linewidth,grid=false]{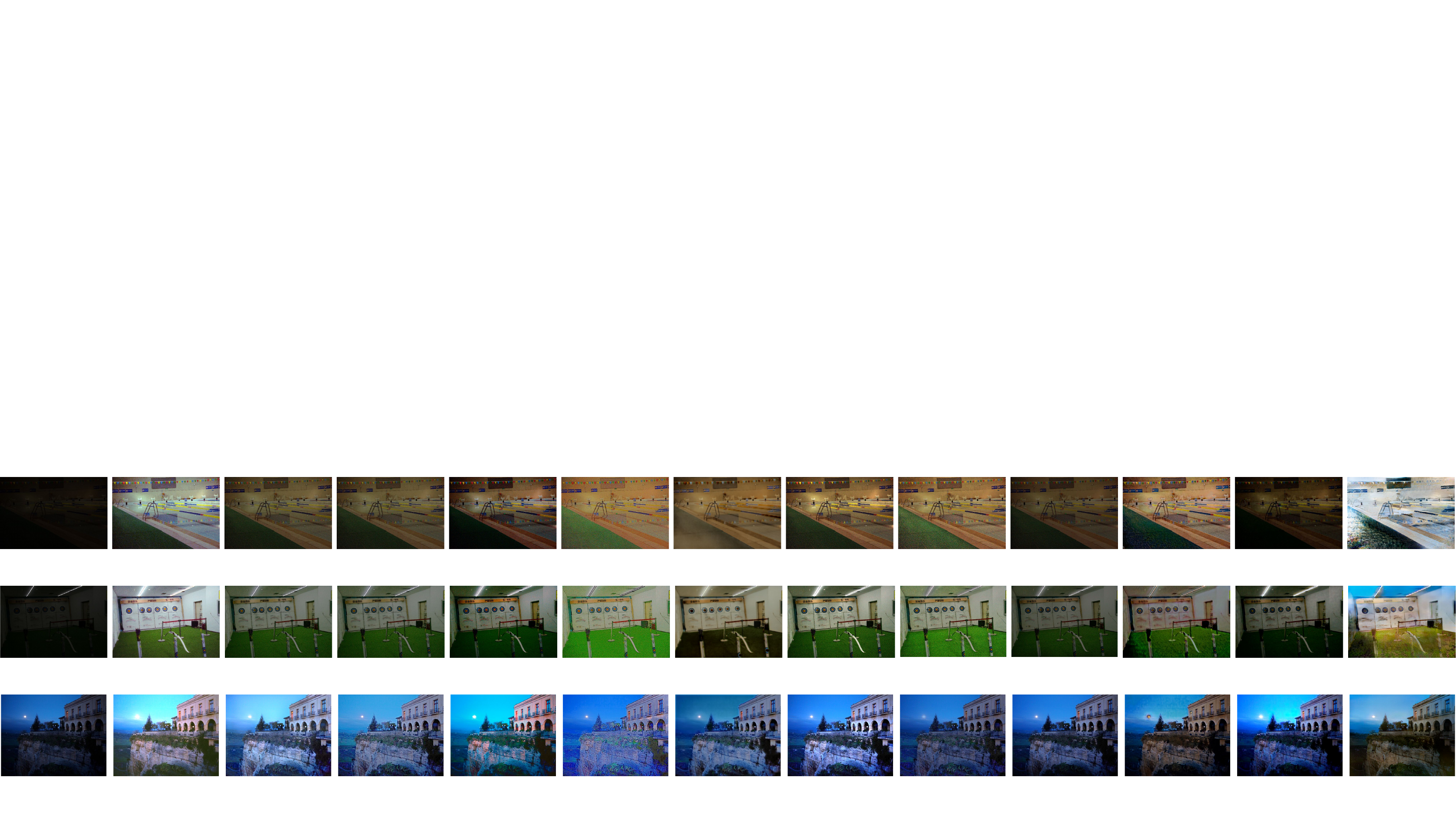}
    \put(2,25){\scriptsize Input}
    \put(10,25){\scriptsize \textbf{Ours}}
    \put(15,25){\scriptsize EnlightenGAN}
    \put(24,25){\scriptsize Zero-DCE}
    \put(32.5,25){\scriptsize CSRNet}
    \put(39,25){\scriptsize RetinexNet}
    \put(48.5,25){\scriptsize DRBN}
    \put(56,25){\scriptsize LIME}
    \put(64,25){\scriptsize NPE}
    \put(71.5,25){\scriptsize SRIE}
    \put(78,25){\scriptsize CycleGAN}
    \put(86.5,25){\scriptsize MIRNet}
    \put(94.5,25){\scriptsize TSIT}

    \put(1,17.5){\tiny LOL-\#690}
    \put(10,17.5){\tiny 4.2425}
    \put(17.75,17.5){\tiny 4.8526}
    \put(25.5,17.5){\tiny 8.3080}
    \put(33.25,17.5){\tiny 9.0266}
    \put(41,17.5){\tiny 8.7709}
    \put(48.75,17.5){\tiny 4.7624}
    \put(56.5,17.5){\tiny 9.2178}
    \put(64.25,17.5){\tiny 9.1257}
    \put(71.75,17.5){\tiny 8.0100}
    \put(78.75,17.5){\tiny 3.6575}
    \put(87,17.5){\tiny 4.9947}
    \put(95,17.5){\tiny 5.7891}    

    \put(1,9.9){\tiny LOL-\#752}
    \put(10,9.9){\tiny 4.3532}
    \put(17.5,9.9){\tiny 5.4467}
    \put(25.5,9.9){\tiny 8.8986}
    \put(33.25,9.9){\tiny 9.5027}
    \put(41,9.9){\tiny 10.216}
    \put(48.75,9.9){\tiny 4.7690}
    \put(56.5,9.9){\tiny 9.3326}
    \put(64.25,9.9){\tiny 9.3576}
    \put(71.75,9.9){\tiny 7.3177}
    \put(78.75,9.9){\tiny 4.9174}
    \put(87,9.9){\tiny 3.9684}
    \put(95,9.9){\tiny 4.6315}
    
    \put(1,2){\tiny LIME-\#2}
    \put(10,2){\tiny 3.1695}
    \put(17.5,2){\tiny 2.0699}
    \put(25.5,2){\tiny 2.1353}
    \put(33.25,2){\tiny 2.4681}
    \put(41,2){\tiny 3.0647}
    \put(48.75,2){\tiny 2.1000}
    \put(56.5,2){\tiny 2.3746}
    \put(64.25,2){\tiny 2.1736}
    \put(71.75,2){\tiny 2.1085}
    \put(78.75,2){\tiny 3.4175}
    \put(87,2){\tiny 2.0576}
    \put(95,2){\tiny 5.3264}
    \end{overpic}
    \vspace{-20pt}
    \caption{Qualitative comparison. We compare out method with the state-of-the-art baselines. The top two rows show results on the images from the test set, while the last row (and also Fig.~\ref{fig:res_eg1}) show results on the LIME dataset~\protect\cite{guo2018lime}. We can see that our method has better generalization ability. Below each image, we also report the NIQE metric for each image. Best viewed by zooming in the electronic version.}
    \label{fig:res_eg2}
\end{figure*}

\paragraph{Spatial Consistency loss.} The spatial consistency loss Eq.~\eqref{eq:spa} can help the network to better learn and infer the normal-light image space explicitly where the spatial coherence between to normal-light images are promoted via this loss. We can observe the contribution of this loss in the third column of Fig.~\ref{fig:ablation}.

\paragraph{Color Consistency loss.} The color consistency loss Eq.~\eqref{eq:color} ensures that the color distribution of the output image does not deviate too much from the input image though the light condition get enhanced significantly. Removing this loss can lead to results with undesirable color distribution (see the fourth column in Fig.~\ref{fig:ablation} for an example).


\subsection{Benchmark Evaluations}
We compare our conditional GAN with several state-of-the-art methods, including SIRE~\cite{fu2016weighted}, LIME~\cite{guo2018lime}, NPE~\cite{wang2013naturalness}, RetinexNet~\cite{deepretinex}, DRBN~\cite{yang2020fidelity}, CSRNet~\cite{he2020conditional}, EnlightenGAN~\cite{jiang2019enlightengan}, Zero-DCE~\cite{guo2020zero}, CycleGAN~\cite{iccv2017cyclegan}, TSIT~\cite{eccv2020tsit}, and MIRNet~\cite{eccv2020mir}. 

Specifically, Table~\ref{tb:psnr_ssim} shows a quantitative comparison between our method and the other baselines on the PSNR, SSIM and the NIQE~\cite{mittal2012niqe} metrics. Note that, even in a more challenging setup without paired information and conditional constraint, our method achieves the state-of-the-art performance on FiveK among unpaired methods. Fig.~\ref{fig:res_eg2} shows a qualitative comparison. In LOL-\#690, we can see our results have brightness and contrast with less noise, and we have enhanced the color and contrast of the trees and the building in LIME-\#2.

\section{Conclusion}
In this paper, we propose a conditional GAN with tailored components including PSM and CCM for image enhancement in an unpaired setting. We also propose task-specific losses including an idempotence loss, a spatial consistency loss, a global color consistency loss, which are combined with the standard GAN loss to encode the reference-controlled channel-wise modulation information and the brightness/contrast information of the input image and the reference image. Our design addresses the one-to-many mapping nature of the problem of image enhancement and achieves state-of-the-art performance on several standard datasets in a much lighter design with 10$\times$ less parameters as compared to the state-of-the-art~\cite{jiang2019enlightengan}. We have justified the usefulness of our designed losses in image enhancement and we believe they can be applied to other image processing tasks such as style transfer or image synthesis due to the fact that our losses encode the general and global information of input images. Therefore, in the future work, we would like to investigate the generalization ability and effectiveness of our designed losses and investigate the performance of our PSM/CCM components in other network architecture.

\section*{Acknowledgments}
We thank Jing Ren for useful suggestions on the manuscript and all the anonymous reviewers for their valuable comments.
\bibliographystyle{named}
\bibliography{ijcai21}

\begin{thebibliography}{}

\bibitem[\protect\citeauthoryear{Brock \bgroup \em et al.\egroup
  }{2018}]{brock2018large}
Andrew Brock, Jeff Donahue, and Karen Simonyan.
\newblock Large scale gan training for high fidelity natural image synthesis.
\newblock In {\em ICLR}, 2018.

\bibitem[\protect\citeauthoryear{Bychkovsky \bgroup \em et al.\egroup
  }{2011}]{fivek2011}
Vladimir Bychkovsky, Sylvain Paris, Eric Chan, and Fr{\'e}do Durand.
\newblock Learning photographic global tonal adjustment with a database of
  input/output image pairs.
\newblock In {\em CVPR}, 2011.

\bibitem[\protect\citeauthoryear{Chen \bgroup \em et al.\egroup
  }{2018a}]{chen2018learning}
Chen Chen, Qifeng Chen, Jia Xu, and Vladlen Koltun.
\newblock Learning to see in the dark.
\newblock In {\em CVPR}, 2018.

\bibitem[\protect\citeauthoryear{Chen \bgroup \em et al.\egroup
  }{2018b}]{chen2018self}
Ting Chen, Mario Lucic, Neil Houlsby, and Sylvain Gelly.
\newblock On self modulation for generative adversarial networks.
\newblock In {\em ICLR}, 2018.

\bibitem[\protect\citeauthoryear{Dumoulin \bgroup \em et al.\egroup
  }{2017}]{dumoulin2016learned}
Vincent Dumoulin, Jonathon Shlens, and Manjunath Kudlur.
\newblock A learned representation for artistic style.
\newblock In {\em ICLR}, 2017.

\bibitem[\protect\citeauthoryear{Fu \bgroup \em et al.\egroup
  }{2016}]{fu2016weighted}
Xueyang Fu, Delu Zeng, Yue Huang, Xiao-Ping Zhang, and Xinghao Ding.
\newblock A weighted variational model for simultaneous reflectance and
  illumination estimation.
\newblock In {\em CVPR}, 2016.

\bibitem[\protect\citeauthoryear{Gharbi \bgroup \em et al.\egroup
  }{2017}]{gharbi2017deep}
Micha\"{e}l Gharbi, Jiawen Chen, Jonathan~T. Barron, Samuel~W. Hasinoff, and
  Fr\'{e}do Durand.
\newblock Deep bilateral learning for real-time image enhancement.
\newblock {\em ACM TOG}, 2017.

\bibitem[\protect\citeauthoryear{Goodfellow \bgroup \em et al.\egroup
  }{2014}]{goodfellow2014generative}
Ian Goodfellow, Jean Pouget-Abadie, Mehdi Mirza, Bing Xu, David Warde-Farley,
  Sherjil Ozair, Aaron Courville, and Yoshua Bengio.
\newblock Generative adversarial nets.
\newblock {\em NeurIPS}, 27, 2014.

\bibitem[\protect\citeauthoryear{Guo \bgroup \em et al.\egroup
  }{2018}]{guo2018lime}
Xiaojie Guo, Yu~Li, and Haibin Ling.
\newblock Lime: Low-light image enhancement via illumination map estimation.
\newblock {\em IEEE TIP}, 2018.

\bibitem[\protect\citeauthoryear{Guo \bgroup \em et al.\egroup
  }{2020}]{guo2020zero}
Chunle Guo, Chongyi Li, Jichang Guo, Chen~Change Loy, Junhui Hou, Sam Kwong,
  and Runmin Cong.
\newblock Zero-reference deep curve estimation for low-light image enhancement.
\newblock In {\em CVPR}, 2020.

\bibitem[\protect\citeauthoryear{He \bgroup \em et al.\egroup
  }{2020}]{he2020conditional}
Jingwen He, Yihao Liu, Yu~Qiao, and Chao Dong.
\newblock Conditional sequential modulation for efficient global image
  retouching.
\newblock In {\em ECCV}, 2020.

\bibitem[\protect\citeauthoryear{Huang and Belongie}{2017}]{huang2017arbitrary}
Xun Huang and Serge Belongie.
\newblock Arbitrary style transfer in real-time with adaptive instance
  normalization.
\newblock In {\em ICCV}, 2017.

\bibitem[\protect\citeauthoryear{Isola \bgroup \em et al.\egroup
  }{2017}]{isola2017image}
Phillip Isola, Jun-Yan Zhu, Tinghui Zhou, and Alexei~A Efros.
\newblock Image-to-image translation with conditional adversarial networks.
\newblock In {\em CVPR}, 2017.

\bibitem[\protect\citeauthoryear{Jiang \bgroup \em et al.\egroup
  }{2020}]{eccv2020tsit}
Liming Jiang, Changxu Zhang, Mingyang Huang, Chunxiao Liu, Jianping Shi, and
  Chen~Change Loy.
\newblock Tsit: A simple and versatile framework for image-to-image
  translation.
\newblock In {\em ECCV}, 2020.

\bibitem[\protect\citeauthoryear{Jiang \bgroup \em et al.\egroup
  }{2021}]{jiang2019enlightengan}
Yifan Jiang, Xinyu Gong, Ding Liu, Yu~Cheng, Chen Fang, Xiaohui Shen, Jianchao
  Yang, Pan Zhou, and Zhangyang Wang.
\newblock Enlightengan: Deep light enhancement without paired supervision.
\newblock {\em IEEE TIP}, 2021.

\bibitem[\protect\citeauthoryear{Land}{1977}]{retinex}
Edwin~H Land.
\newblock The retinex theory of color vision.
\newblock {\em Scientific american}, 1977.

\bibitem[\protect\citeauthoryear{Loh and Chan}{2019}]{exdark2019}
Yuen~Peng Loh and Chee~Seng Chan.
\newblock Getting to know low-light images with the exclusively dark dataset.
\newblock {\em CVIU}, 2019.

\bibitem[\protect\citeauthoryear{Lore \bgroup \em et al.\egroup }{2017}]{llnet}
Kin~Gwn Lore, Adedotun Akintayo, and Soumik Sarkar.
\newblock Llnet: A deep autoencoder approach to natural low-light image
  enhancement.
\newblock {\em Pattern Recognition}, 2017.

\bibitem[\protect\citeauthoryear{Mirza and
  Osindero}{2014}]{mirza2014conditional}
Mehdi Mirza and Simon Osindero.
\newblock Conditional generative adversarial nets.
\newblock {\em arXiv:1411.1784}, 2014.

\bibitem[\protect\citeauthoryear{Mittal \bgroup \em et al.\egroup
  }{2012}]{mittal2012niqe}
Anish Mittal, Rajiv Soundararajan, and Alan~C. Bovik.
\newblock Making a completely blind image quality analyzer.
\newblock {\em IEEE SPL}, 2012.

\bibitem[\protect\citeauthoryear{Park \bgroup \em et al.\egroup
  }{2019}]{park2019semantic}
Taesung Park, Ming-Yu Liu, Ting-Chun Wang, and Jun-Yan Zhu.
\newblock Semantic image synthesis with spatially-adaptive normalization.
\newblock In {\em CVPR}, 2019.

\bibitem[\protect\citeauthoryear{Pizer \bgroup \em et al.\egroup }{1987}]{HE}
Stephen~M Pizer, E.~Philip Amburn, John~D Austin, Robert Cromartie, Ari
  Geselowitz, Trey Greer, Bart ter Haar~Romeny, John~B Zimmerman, and Karel
  Zuiderveld.
\newblock Adaptive histogram equalization and its variations.
\newblock {\em CVGIP}, 1987.

\bibitem[\protect\citeauthoryear{Reed \bgroup \em et al.\egroup
  }{2016}]{reed2016generative}
Scott Reed, Zeynep Akata, Xinchen Yan, Lajanugen Logeswaran, Bernt Schiele, and
  Honglak Lee.
\newblock Generative adversarial text to image synthesis.
\newblock In {\em ICML}, 2016.

\bibitem[\protect\citeauthoryear{Ronneberger \bgroup \em et al.\egroup
  }{2015}]{ronneberger2015u}
Olaf Ronneberger, Philipp Fischer, and Thomas Brox.
\newblock U-net: Convolutional networks for biomedical image segmentation.
\newblock In {\em MICCAI}, 2015.

\bibitem[\protect\citeauthoryear{Wang \bgroup \em et al.\egroup
  }{2013}]{wang2013naturalness}
Shuhang Wang, Jin Zheng, Hai-Miao Hu, and Bo~Li.
\newblock Naturalness preserved enhancement algorithm for non-uniform
  illumination images.
\newblock {\em IEEE TIP}, 2013.

\bibitem[\protect\citeauthoryear{Wang \bgroup \em et al.\egroup
  }{2018}]{wang2018high}
Ting-Chun Wang, Ming-Yu Liu, Jun-Yan Zhu, Andrew Tao, Jan Kautz, and Bryan
  Catanzaro.
\newblock High-resolution image synthesis and semantic manipulation with
  conditional gans.
\newblock In {\em CVPR}, 2018.

\bibitem[\protect\citeauthoryear{Wei \bgroup \em et al.\egroup
  }{2018}]{deepretinex}
Chen Wei, Wenjing Wang, Wenhan Yang, and Jiaying Liu.
\newblock Deep retinex decomposition for low-light enhancement.
\newblock In {\em BMVC}, 2018.

\bibitem[\protect\citeauthoryear{Yang \bgroup \em et al.\egroup
  }{2020}]{yang2020fidelity}
Wenhan Yang, Shiqi Wang, Yuming Fang, Yue Wang, and Jiaying Liu.
\newblock From fidelity to perceptual quality: A semi-supervised approach for
  low-light image enhancement.
\newblock In {\em CVPR}, 2020.

\bibitem[\protect\citeauthoryear{Zamir \bgroup \em et al.\egroup
  }{2020}]{eccv2020mir}
Syed~Waqas Zamir, Aditya Arora, Salman Khan, Munawar Hayat, Fahad~Shahbaz Khan,
  Ming-Hsuan Yang, and Ling Shao.
\newblock Learning enriched features for real image restoration and
  enhancement.
\newblock In {\em ECCV}, 2020.

\bibitem[\protect\citeauthoryear{Zhang \bgroup \em et al.\egroup
  }{2020}]{zhang2020cross}
Pan Zhang, Bo~Zhang, Dong Chen, Lu~Yuan, and Fang Wen.
\newblock Cross-domain correspondence learning for exemplar-based image
  translation.
\newblock In {\em CVPR}, 2020.

\bibitem[\protect\citeauthoryear{{Zhu} \bgroup \em et al.\egroup
  }{2017}]{iccv2017cyclegan}
Jun-Yan {Zhu}, Taesung {Park}, Phillip {Isola}, and Alexei~A. {Efros}.
\newblock Unpaired image-to-image translation using cycle-consistent
  adversarial networks.
\newblock In {\em ICCV}, 2017.

\end{thebibliography}

\end{document}